\documentclass{article}

\usepackage{microtype}
\usepackage{graphicx}
\usepackage{subfigure}
\usepackage{booktabs} % for professional tables
\usepackage{amsfonts}
\usepackage{amsmath}
\usepackage{mathtools}
\usepackage{amsthm}

\newtheorem{lemma}{Lemma}
\newtheorem{theorem}{Theorem}
\newtheorem{corollary}{Corollary}
\theoremstyle{definition}

\newcommand\numberthis{\addtocounter{equation}{1}\tag{\theequation}}

\usepackage{hyperref}

\usepackage[accepted]{icml2020}

\icmltitlerunning{Neural Inverse Transform Sampler}

\begin{document}

\twocolumn[
\icmltitle{Neural Inverse Transform Sampler}



\icmlsetsymbol{equal}{*}

\begin{icmlauthorlist}
\icmlauthor{Henry Li}{yale}
\icmlauthor{Yuval Kluger}{yale}
\end{icmlauthorlist}

\icmlaffiliation{yale}{Department of Applied Mathematics, Yale University, New Haven, CT}

\icmlcorrespondingauthor{Henry Li}{henry.li@yale.edu}

\icmlkeywords{Machine Learning, ICML}

\vskip 0.3in
]




\printAffiliationsAndNotice{}

\begin{abstract}
Any explicit functional representation $f$ of a density is hampered by two main obstacles when we wish to use it as a generative model: designing $f$ so that sampling is fast, and estimating $Z = \int f$ so that $Z^{-1}f$ integrates to 1. This becomes increasingly complicated as $f$ itself becomes complicated. In this paper, we show that when modeling one-dimensional conditional densities with a neural network, $Z$ can be exactly and efficiently computed by letting the network represent the cumulative distribution function of a target density, and applying a generalized fundamental theorem of calculus. We also derive a fast algorithm for sampling from the resulting representation by the inverse transform method. 
% These ideas form a basic probabilistic unit, which can then be extended to higher dimensions.
By extending these principles to higher dimensions, we introduce the \textbf{Neural Inverse Transform Sampler (NITS)}, a novel deep learning framework for modeling and sampling from general, multidimensional, compactly-supported probability densities. NITS is a highly expressive density estimator that boasts end-to-end differentiability, fast sampling, and exact and cheap likelihood evaluation. We demonstrate the applicability of NITS by applying it to realistic, high-dimensional density estimation tasks: likelihood-based generative modeling on the CIFAR-10 dataset, and density estimation on the UCI suite of benchmark datasets, where NITS produces compelling results rivaling or surpassing the state of the art.

\end{abstract}

\begin{figure*}[ht]
\vskip -0.4in
\begin{center}
% \centerline{\includegraphics[width=\columnwidth]{pixelcnn_ours}}
\centering
\subfigure{\label{fig:a}\includegraphics[width=160mm]{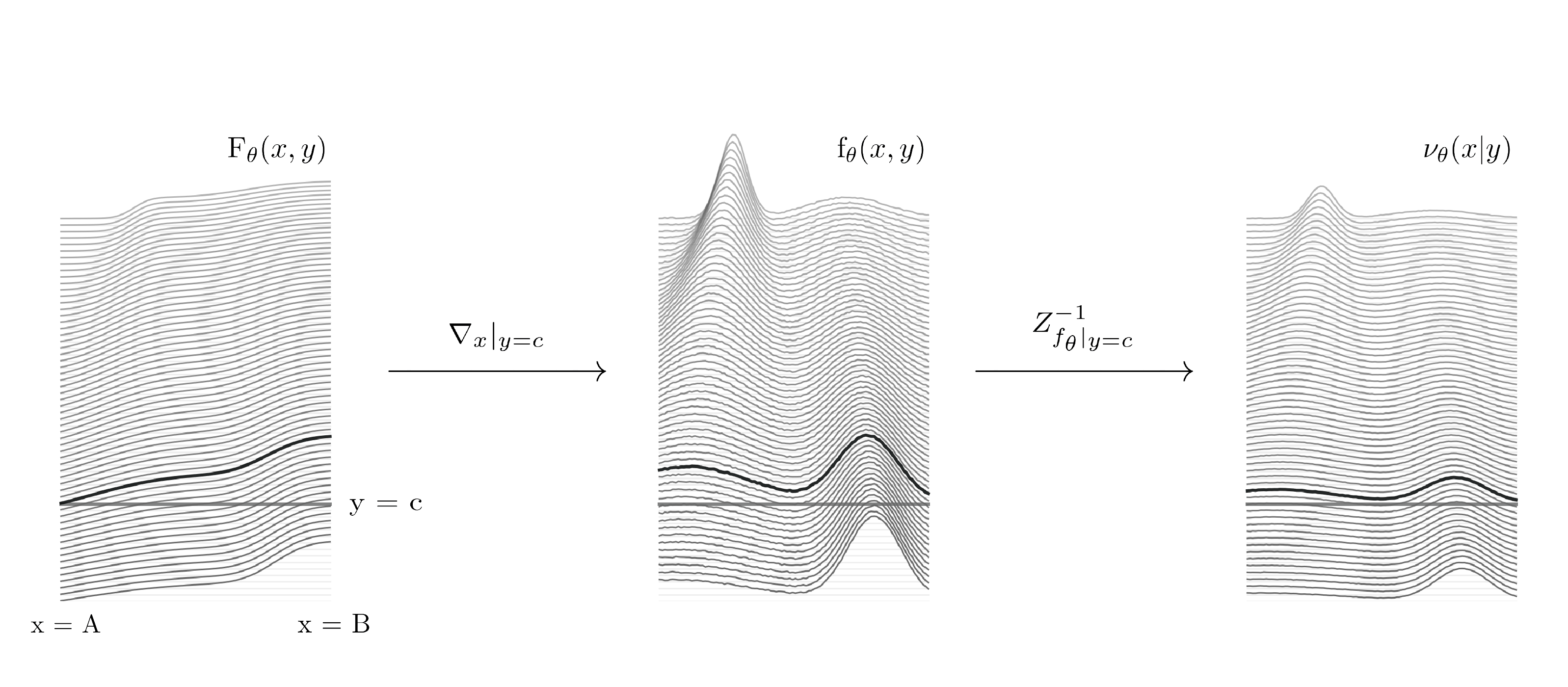}}

\caption{The \textit{integration trick}, used here to enable direct representation of a conditional density $\nu_\theta(x|y=c)$. The original neural network (left) is differentiated w.r.t. $x$ (middle), then rescaled by the partition function (right), which we compute directly by evaluating $F_\theta$ at the boundaries of $[A, B]$. All operations are restricted to the line $y=c$ (for more details, see Sections \ref{modeling_1d} and \ref{modeling_multi_d}).}
\label{fig:pnn}
\end{center}
\vskip -0.2in
\end{figure*}

\section{Introduction}
\label{submission}

Building efficient, highly expressive generative density estimators has been a long-standing challenge in machine learning. A sufficiently powerful estimator will have far reaching effects on a bevy of inference tasks, ranging from classification to generative modeling to missing value imputation. 

Density estimation of continuous random variables revolves around a fundamental trade off between expressivity and tractability, which is most concretely related to the computation of the \textit{partition function} $Z_\theta = \int_\Omega f_\theta$ for a positive function $f_\theta: \Omega \rightarrow \mathbb{R}_+$. The general rule of thumb: the more flexible the underlying $f_\theta$, the more difficult the estimation of $Z_\theta$.

For instance, the most general model is perhaps the energy-based model (EBM), where the desired density $\nu$ is approximated via an energy landscape  --- for example, the Boltzmann (or Gibbs) distribution
\begin{equation}
    \nu_\theta = \frac{e^{-f_\theta(x)}}{\int_\Omega e^{-f_\theta(x)} dx},
\end{equation}
where $f_\theta$ may be an arbitrary function. The problem with this formulation is that analytic integration of functions of general form is unknown, and the cost of numerical integration techniques scales exponentially with the dimension of $\Omega$. Thus there are few, if any, approaches that represent an arbitrary $f_\theta$ in high dimension, though some methods may estimate $f_\theta$ via a variational lower bound \cite{sohl2015deep,du2019implicit}.

On the other hand, one may obtain $Z_\theta$ through direct analytic derivation --- but this is only applicable for a small handful of distributions. \cite{theis2012mixtures} and \cite{uria2016neural}, for example, perform density estimation via Gaussian mixture and Gaussian scale mixture models, which both have well-known partition functions.

Ultimately, most approaches avoid explicit estimation of the density. Normalizing flows \cite{dinh2014nice, rezende2015variational} learn the transform between the desired density and a reference density (usually Gaussian), and compute $\nu$ via the probabilistic change-of-variables formula, which does not require $Z$, while score-based \cite{song2020score} models work with the gradient of the log-likelihood, which again does not depend on the partition function. These techniques for sidestepping the computation of $Z$ each have drawbacks, which we will explore in Section \ref{related_work}.

Directly approximating a pdf $\nu$ with a parametric function $f_\theta$ has remained an elusive task. This is also the approach we take on with the proposed Neural Inverse Transform Sampler (NITS). NITS is a deep neural network augmented by two basic ideas: fast integration via the \textbf{gradient theorem}, and fast sampling via the \textbf{inverse transform method}. 

In the one-dimensional case, we leverage the fundamental theorem of calculus to normalize a neural network's output $F_\theta(x)$ so that it is the cumulative distribution function (cdf) of a 1D probability distribution over a bounded interval $[A, B]$. This also enforces the neural network derivative $\nabla_x F_\theta(x)$ to be the corresponding probability density function (pdf). Fast evaluation of the cdf means fast sampling from the induced distribution using the inverse transform method. Furthermore, the parameters of the distribution can be trained via gradient descent, as the framework is end-to-end differentiable.

This idea is straightforwardly extended to higher dimensions, where a multi-dimensional probabilistic model is composed of multiple one-dimensional models whose statistics are computed either in parallel or in sequence, according to assumptions on the correlation structure of the random variables. See Figure \ref{fig:pnn} for a graphical breakdown. In Section \ref{background} we explore these ideas in greater detail.

\textbf{Our Contributions} \hspace{.05in} In this work, we demonstrate two novel computational ideas: a method for obtaining cheap and exact integrals of 1D neural network functions which we call the \textit{integration trick}, and an architecture for density estimation that allows for fast and accurate sampling via the inverse transform method. These techniques enable the design of a deep learning framework for generalized density estimation of multi dimensional, compactly supported, continuous valued random variables, which we call the Neural Inverse Transform Sampler\footnote{Code available at \href{https://github.com/lihenryhfl/NITS}{github.com/lihenryhfl/NITS}.}. NITS is (to our knowledge) the first framework to provide explicit\footnote{Explicit, in the sense that the density is directly represented by a neural network.} representations for densities with non-analytical partition functions.

We apply the resulting framework in two scenarios. First, in a generative autoregressive density model on natural images. And second on the UCI suite of density estimation benchmark datasets. We report competitive results in both settings.

Finally, we corroborate our empirical results by demonstrating that our model can universally approximate compactly supported densities of continuous random variables.

\section{Probabilistically Normalized Networks}
\label{background}
We consider the task of representing a parametric family of compactly supported probability densities on $[A, B]^n \subset \mathbb{R}^n$, $A,B \in \mathbb{R}$.
An ideal model should possess the following properties: (1) high expressiveness, (2) fast sampling, (3) fast computation of $\nu_\theta$, and (4) end-to-end differentiability. 

% For notation, we will let $\mathcal{P}$ denote the parametric family, $P_\theta$ denote an instance of $\mathcal{P}_\theta$, $\nu_\theta$ denote its density, and $X \sim P_\theta$ denote a random variable sampled from $P_\theta$.

In this section we propose the \textbf{probabilistically normalized network} (PNN), which induces such a family. We will start from the one dimensional case, introducing the modeling framework in Section \ref{modeling_1d}, and the sampling framework in section \ref{sampling_1d}. We will then generalize to the higher dimensional case in Section \ref{modeling_multi_d} in two ways: first, assuming that $X = (X_1, \dots, X_n)^T$ are independent, and second, that they are autoregressively correlated. 
% In Section \ref{inverse_pnn} we will explore the differentiability of the sampling algorithm.

% After defining the, in Section \ref{expressiveness} we will then show that the PNN can universally represent the class of all distributions under their respective assumptions.

\subsection{Modeling a Compactly Supported 1D Distribution}
\label{modeling_1d}

We begin with a continuous compactly supported one dimensional random variable $X \sim \nu_\theta$. We would like to express $\nu_\theta: [A, B] \rightarrow \mathbb{R}$ as some neural network function parameterized by $\theta$. To be a valid density, $\nu_\theta$ must satisfy two conditions.

\textbf{Condition 1: Integration to Unity} \hspace{.05in} i.e.,
\begin{equation}
    \int_{A}^B \nu_\theta(x) dx = 1. \label{eq:unity}
\end{equation}
For this to hold, we choose to form the decomposition $\nu_\theta = f_\theta / Z_\theta$, where $f_\theta$ can now be an arbitrary neural network, and $Z_\theta = \int_\mathbb{R} f_\theta(x) dx$ is the partition function that normalizes $f_\theta$ so that the resulting $\nu_\theta$ will integrate to $1$.

Unfortunately, when $f_\theta$ is a simple neural network, any integral including $Z_\theta$ can only be computed via numerical quadrature (e.g., neural ODE frameworks  \cite{chen2018neural,wehenkel2019unconstrained}), which is expensive in low dimensions and virtually intractable in higher dimensions.

However, by letting $f_\theta$ be the \textit{derivative} of a neural network, we can compute $\nu$ in $\mathcal{O}(1)$ time (with respect to the size of the integration region). To do this, we leverage the crucial observation that, in one dimension, whenever there exists another function $F_\theta: X \rightarrow \mathbb{R}$ such that
\begin{equation}
    \label{eq:antiderivative}
    F_\theta(x) = \frac{d}{dx} f_\theta(x) \hspace{.1in} \text{for all $x$ in $\mathbb{R}$},
\end{equation} 
the integral of $f_\theta$ over any bounded interval $[a,b]$ can be evaluated via the formula
\begin{equation}
    \int_a^b f_\theta(t) dt = F_\theta(b) - F_\theta(a),
\end{equation}
due to the \textbf{fundamental theorem of calculus}.

Consequently, when $F_\theta(x)$ is a neural network and $f_\theta(x) := \nabla_x F_\theta(x)$ its derivative, the required assumption Eq. (\ref{eq:antiderivative}) is automatically satisfied, and the pdf $\nu_\theta$ becomes computable via
\begin{equation}
    \label{eq:pdf}
    \nu_\theta(x) = \frac{f_\theta(x)}{F_\theta(B) - F_\theta(A)}.
\end{equation}
As the denominator is $Z_\theta = \int_\mathbb{R} f_\theta(x) dx = \int_{A}^B f_\theta(x) dx$, Condition 1 (i.e. Eq. \ref{eq:unity}) must hold. This technique, which we term the \textit{integration trick}, is cheap, exact, and preserves gradients of $F_\theta$ and $f_\theta$ with respect to both $\theta$ and $x$.

Moreover, it is now clear that $F_\theta$ divided by the same denominator as in Eq. \ref{eq:pdf} must be the cdf corresponding to the density $\nu_\theta$:
\begin{equation}
    \label{eq:cdf}
    N_\theta(x) = \frac{F_\theta(x)}{F_\theta(B) - F_\theta(A)} - F_\theta(A),
\end{equation}
where we subtract by $F_\theta(A)$ so that $N_\theta(A) = 0$. Additionally, now that we have defined $N_\theta(x)$, we note that we can obtain $\nu_\theta$ directly via $\nu_\theta(x) = \nabla_x N_\theta(x)$ (we previously obtained $\nu_\theta$ by scaling the gradient of $F_\theta$). This is because the differentiation and rescaling operators commute.

\textbf{Condition 2: Positivity} \hspace{.05in} i.e.,
\begin{equation}
    \nu_\theta(x) > 0 \hspace{.05in} \text{for all $x$ in $[A, B]$}.
\end{equation}
This is equivalent to requiring that \begin{equation}
    \label{eq:positive_grad}
    \nabla_x F_\theta \geq 0 \hspace{.1in} \text{for all $x$ in $[A, B]$},
\end{equation}
as $\nu_\theta$ is proportional to $\nabla_x F_\theta$ up to a factor $Z_\theta$, which is always positive. There are many ways of enforcing Eq. \ref{eq:positive_grad} --- we take the simple approach of using positive monotonic activations (e.g. sigmoid, tanh, ReLU) and enforcing the positivity of the weights of the neural network $F_\theta$. This is sufficient for Condition 2, given the following lemma:

\begin{lemma}
\label{thm:monotonic}
A fully connected neural network $F_\theta: \mathbb{R}^n \rightarrow \mathbb{R}$ with positive weights and positive monotonic activations has directional derivative $f_\theta := \nabla_{x_i} F_\theta > 0$ for all $i = 1, \dots n$.
\end{lemma} 

\textbf{Probabilistically Normalized Network} \hspace{.05in} As both $\nu_\theta$ and $N_\theta(x)$ describe a rescaled neural network (scaled by the same constant $Z_\theta^{-1} = (F_\theta(B) - F_\theta(A))^{-1}$), we call the resulting function a probabilistically normalized network, or PNN.

\subsection{Sampling from a Compactly Supported 1D Distribution}
\label{sampling_1d}
Sampling from the PNN-induced distribution is now straightforward given our cheap access to $N_\theta$. We can use the well-known inverse transform theorem, which we state below for completeness.
\begin{lemma}
\label{thm:itm}
(\textbf{Inverse Transform Theorem}) Let $N_\theta$ be as defined above, and $N^{-1}_\theta(y)$, $y \in [0,1]$ denote its inverse, i.e.
\begin{equation}
    \label{eq:inverse}
    N_\theta^{-1}(y) = \min\{x : N_\theta(x) \geq y\} \hspace{.1in} y \in [0, 1].
\end{equation}
If we let $\mu$ be the Lebesgue density on $[0, 1]$, then $\nu_\theta$ is its pushforward density via the transform $N_\theta^{-1}$, i.e.
\begin{equation}
    \nu_\theta = N_{\theta\ast}^{-1} (\mu).
\end{equation}
\end{lemma}
Thus, sampling from $x \sim \nu_\theta$ is a simple two step process: 
\begin{enumerate}
    \item draw $z \sim \text{Unif}[0,1]$,
    \item compute $x = N_\theta^{-1}(z)$,
\end{enumerate}
where $N_\theta^{-1}$ can be accurately and efficiently calculated via bisection search, due to the monotonic nature of the cdf $N_\theta$.

\subsection{Modeling Multi-Dimensional Distributions}
\label{modeling_multi_d}
Let us now consider the case of modeling $X = (X_1, ..., X_n)^T$, an $n > 1$ dimensional, continuous, compactly supported random variable. By making various further assumptions on the correlation structure of the coordinates in $X$, we can straightforwardly extend the 1D formulation in the previous subsection to higher dimensions. Here we consider two cases: a coordinate-wise independent correlation structure, and an autoregressive correlation structure.

\textbf{Coordinate-wise Independent Model} \hspace{.05in} Suppose that all elements of $X$ are independently but perhaps not identically distributed. Note that, while simple, this correlation structure is not supported by normalizing flow-based approaches \cite{dinh2014nice, rezende2015variational} due to their invertibility constraint, which by design \textit{requires} dependence between coordinates. Let $N_\theta: \mathbb{R}^n \rightarrow [0, 1]^n$ be the concatenation of $n$ PNNs (and $\theta$ be the concatenation of their parameters), i.e.,
\begin{equation}
    N_\theta = (N_{\theta_1}, \dots, N_{\theta_n})^T \hspace{.1in} \text{and} \hspace{.1in} \theta = (\theta_1, \dots, \theta_n)^T.
\end{equation}
Then the density of the resulting distribution is simply the trace of its Jacobian
\begin{equation*}
    \nu_\theta(x_1, \dots, x_n) = \text{tr}(J_{N_{\theta}}) = \prod_{i=1}^n \nu_{\theta_i}(x_i),
\end{equation*}
where $J_f$ denotes the Jacobian of $f$. And since the per-dimension cdf is a 1D PNN, sampling only requires a bisection search as before, and can be performed over all dimensions in parallel.

\textbf{Autoregressive Model} \hspace{.05in} Retaining the previous definition of $X$, we consider the case where $X_i$ depends on $X_j$ for all $j < i$. This requires the PNN $F_{\theta_i}$ to take multi-dimensional inputs, i.e. $F_{\theta_i}: \mathbb{R}^i \rightarrow [0, 1]$, as we are now modeling the conditional likelihood $P(X_i | X_{<i})$. Note that the parameters $\theta_i$ can be different for each $i$. 

To obtain the conditional variants of $N_\theta$ and $\nu_\theta$, we apply the following theorem, which is a multi-dimensional generalization of the fundamental theorem of calculus, reproduced below for completeness.
\begin{lemma}
\label{thm:gt}
(\textbf{Gradient Theorem}) Let $F: \mathbb{R}^n \rightarrow \mathbb{R}$ be a continuously differentiable function and $\varphi:[a,b] \rightarrow \mathbb{R}^n$ be a curve in $\mathbb{R}^n$, where $a, b \in \mathbb{R}$ and $\varphi(a), \varphi(b)$ are the endpoints of the curve. Then
\begin{equation}
    \int_{\varphi[a,b]} \nabla F \cdot dr = F(\varphi(b)) - F(\varphi(a)).
\end{equation}
\end{lemma}

% Letting $F_{\theta_i}: \mathbb{R}^i \rightarrow \mathbb{R}$ be a neural network and $\varphi_{x_{<i}}(x_i) = x_ie_i + {x_{<i}} \odot (1 - e_i)$ be the curve in $\mathbb{R}^i$, where $e_i$ is the $i$-dimensional one-hot vector taking a nonzero value at coordinate $i$, we can define
% \begin{align*}
%     G_{\theta_i}(x_i, x_{<i}) &:= \int_{\varphi_{x_{<i}}[A, x_i]} \nabla F_{\theta_i} \cdot dr \\
%     &= \int_{A}^{x_i} \frac{d}{dx_i} F_{\theta_i}(x_i, x_{<i}) dx_i \numberthis \label{eq:G_theta}. \\
%     % &= F_{\theta_i}(x_i, \nabla_{x<i})
%     % g_{\theta_i}(x_i, x_{<i}) &:= \frac{d}{dx_i} F_{\theta_i}(x_i, x_{<i}). \numberthis \label{eq:g_theta}
% \end{align*}
% Subsequently, we normalize $G_{\theta_i}(x_i, x_{<i})$ in a manner similar to the 1D case, obtaining
Let $F_{\theta_i}: \mathbb{R}^i \rightarrow \mathbb{R}$ be a neural network and $\varphi_{x_{<i}}(x_i)$ be the line along the $i$-th coordinate direction passing through $x_{<i}$, i.e.,
\begin{equation}
    \varphi_{x_{<i}}(x_i) = x_ie_i + {x_{<i}} \odot (1 - e_i),
\end{equation}
where $\odot$ is the Hadamard product and $e_i$ is the $i$th basis vector. Then a conditional cdf of $x_i$, conditioned on the first $i-1$ coordinates of $x$, may be written as
\begin{align}
    N_{\theta_i}(x_i | x_{<i}) &= \frac{F_{\theta_i}(x_i, x_{<i})}{F_{\theta_i}(B, x_{<i}) - F_{\theta_i}(A, x_{<i})} - F_{\theta_i}(A, x_{<i}) \label{eq:ar_cdf},
\end{align}
where $[A,B]$ contains the compact support of $x_i$.
Again, $\nu_{\theta_i}(x_i | x_{<i})$ can simply be computed from Eq. \ref{eq:ar_cdf} by differentiating w.r.t. $x_i$, i.e.,
\begin{equation*}
    \nu_{\theta_i}(x_i | x_{<i}) = \frac{d}{dx_i} N_{\theta_i}(x_i | x_{<i})
\end{equation*}
See Figure \ref{fig:pnn} for a graphical representation of the rescaling process.

As in the 1D case, we enforce $F_{\theta_i}$ to have positive weights and monotonic activation functions. Letting $\theta = (\theta_1, \dots, \theta_n)^T$, the above definitions produce the joint density
\begin{equation*}
    \nu_\theta(x_1, \dots, x_n) = \prod_{i=1}^n \nu_{\theta_i}(x_i | x_{<i}),
\end{equation*}
and we refer to the underlying rescaled neural network $N_\theta$ as an \textbf{autoregressive PNN}.

To draw points from the induced distribution $\nu_\theta$, we use ancestral sampling, and draw each coordinate via the inverse transform method. Namely, to obtain $x_i \sim \nu_\theta$ given previously sampled $x_{<i}$, we
\begin{enumerate}
    \item sample $z \sim \text{Unif}[0,1]$,
    \item compute $x_i = N_\theta^{-1}(z | x_{<i})$,
\end{enumerate}
where $N_\theta^{-1}(z | x_{<i})$ is the inverse of $N_\theta(x_i | x_{<i})$ with respect to the input $x_i$ (the conditional case of Eq. \ref{eq:inverse}). When $i=1$, $x_1$ is sampled in the same manner as the 1D case, since $N_\theta(x_1)$ is not conditionally dependent on any other coordinates.

\section{Neural Inverse Transform Sampler}
\label{architecture}
Our proposed Neural Inverse Transform Sampler for a general, compactly supported $n$-dimensional random variable $X$ consists of two components:
\begin{enumerate}
    \item a set of concatenated PNN functions $N_\theta = (N_{\theta_1}, \dots, N_{\theta_n})^T$
    \item a corresponding weight model $W_\phi(x)$ that supplies the parameters $\theta$.
\end{enumerate}
In this paper, we focus on maximum likelihood estimation (MLE), and perform density estimation by
\begin{equation}
    \max_\phi \sum_{i=1}^n \log \nu_{W_\phi(x_i)}(x_i),
\end{equation}
where $\nu_\theta(x)$ is the PNN density parameterized by $\theta$ obtained from the weight model $W_\phi$. However, this is not the only way to train a NITS model, as any likelihood-based loss can be used.

NITS has a fast sampling scheme, supports efficient and exact likelihood evaluation, and is expressive and end-to-end differentiable. Algorithm \ref{alg:sampling} describes the process for sampling from $P_\theta$ in the autoregressive (and most general) variant of NITS\footnote{\label{algnote1}In Algorithm \ref{alg:sampling}, $N_\theta^{-1} = \mathtt{monotonic\_inverse}(z, \mathtt{f}(\cdot), \epsilon)$ is a function that takes a scalar $z$, tolerance $\epsilon$, and monotonic function $N_\theta$ and returns $x$ such that $|N_\theta(x) - z| < \epsilon$, i.e., it inverts $N_\theta$. The monotonicity of $N_\theta$ permits the use of bisection search, which has runtime $\mathcal{O}(\log(1/\epsilon))$.}. In the following subsections, we discuss important properties and implementational details. 

\begin{algorithm}[tb]
  \caption{Sampling from $P_\theta$}
  \label{alg:sampling}
\begin{algorithmic}
    \STATE {\bfseries Input:} PNN $N_\theta(x)$, parameter network $W_\phi(x)$
    \STATE Initialize $x[i] = 0$ for $i = 1, \dots, n$.
    \STATE Sample $\mathbf{z}[i] \sim \mathtt{Uniform}[0, 1]$ for $i = 1, \dots, n$.
    \FOR{$i=1$ {\bfseries to} $n$}
    \STATE $\theta_i = W_\phi(x)$
    \STATE $x[i] = N_{\theta_i}^{-1}(z_i | x)\footnotemark[2\ref{algnote1}]$
    \ENDFOR
\end{algorithmic}
\end{algorithm}

\subsection{Architecture}
For a neural network of $n$ layers, our proposed architecture for $F_\theta$ can be specified by a recursive definition. Letting $a_\ell$ denote the activations at the $\ell$th layer of the PNN (and $a_0 = x$),
\begin{equation}
    \label{eq:pnn_architecture}
    a_\ell = \sigma(h_A(A_\ell)^T a_{\ell - 1} + h_b(b_\ell, A_\ell))
\end{equation}
for $\ell = 1, 2, \dots, n - 1$, where $\sigma$ is the sigmoid activation function, and $\{A_\ell, b_\ell\}$ denote the weights and biases of the $\ell$-th layer. The final layer of the PNN is defined differently:
\begin{equation}
    \label{eq:pnn_final_layer}
    F_\theta(x) = a_n = h_s(A_n)^T a_{n-1}.
\end{equation}
$h_A, h_b,$ and $h_s$ all denote parameter transformation functions, which we will subsequently define.

As discussed in Section \ref{modeling_1d}, to obtain the density $\nu_\theta \propto \nabla_x F_\theta$, we required that 1) $\int \nu_\theta(x) dx = 1$, and 2) $\nu_\theta(x) \geq 0$ for all $x$. We enforced condition 1) by dividing the output of $F_\theta$ by its integral over the bounds $[A, B]$, as in Eqs. (\ref{eq:pdf}, \ref{eq:cdf}, and \ref{eq:ar_cdf}). This is cheap and exact by our framework, via application of Lemma \ref{thm:gt}. Condition 2) was enforced by judicious choice of activation function $\sigma$ and weight parameter transformation functions and Lemma \ref{thm:monotonic}:
\begin{align}
    \sigma(x) &= \mathtt{sigmoid}(x) \\
    h_A(A) &= exp(-A),
\end{align}

The final two parameter transformation functions are notable deviations from a regular fully connected neural network: 
\begin{align}
    h_b(b, A) &= -exp(-A) \odot b \\
    h_s(A) &= \mathtt{softmax}(A).
\end{align}
In words, we apply a parameter transformation function $h_b$ to the bias $b$, and a softmax function in the final layer to $A$ (the weights, the activations). 

Note that this parameterization is equivalent to that of a convex sum of standard multilayer perceptrons \footnote{To see this, examine Eq. \ref{eq:pnn_architecture}, and note that choosing $C_\ell = h_A(A_\ell)$ and $d_\ell = h_b(b_\ell, A_\ell)$ produces the recurrence $a_\ell = C_\ell^T a_{\ell - 1} + d_\ell$, which is a standard perceptron formulation. Then Eq. \ref{eq:pnn_final_layer} is a convex combination of such networks.}, and these modifications are not necessary to enforce monotonicity or positivity of the weights. Instead, they serve to bridge the gap between NITS and another parametric family of distributions, the \textbf{mixture of logistics distribution}, which we will explore in the proceeding subsection.

\subsection{NITS is a Deep Mixture of Logistics}
\label{nits_mol}
The mixture of logistics is a common distribution in statistics and probability, with density and cdf
\begin{align}
    f(x) &= \sum_{i=1}^n \alpha_i \frac{e^{(x - \mu_i) / s_i}}{s_i(1 + e^{(x - \mu_i)/s_i})^2} \\
    F(x) &= \sum_{i=1}^n \alpha_i \sigma((x - \mu_i) / s_i), \label{eq:mol_cdf}
\end{align}
respectively, for a mixture of size $n$ and parameters $(\{\mu_i\}_{i=1}^n, \{s_i\}_{i=1}^n)$. 

On the other hand, letting $a_{n-2} = F_\theta^{n-2}(x)$ denote the input to the penultimate layer of $F_\theta$, we observe that the final two layers of $F_\theta$ can be written as:
\begin{align*}
    F_\theta(x) %&= h_s(A_n)^T \sigma(h_A(A_{n-1})^Ta_{n-2} - h_b(b_{n-1}, A_{n-1})) \\
    &= \mathtt{softmax}(A_n)^T \sigma((a_{n-2} - b_{n-1}) / \exp(A_{n-1})) \\
    &= \sum_{i=1}^k \beta_i \sigma((a_{n-2} - b_i) / s_i) \numberthis \label{eq:mol_like_cdf}
\end{align*}
where $k$ is the width of the final hidden layer of $F_\theta$, $\beta_i = \mathtt{softmax}(A_n)_i$, and $s_i = \exp(A_{n-1})_i$.

Comparing Eqs. \ref{eq:mol_cdf} and \ref{eq:mol_like_cdf}, it is easy to see that the cdf $N_\theta$ of a two-layer PNN would be nearly identical to a mixture of logistics distribution, with the exception of the normalization by $Z_\theta^{-1}$ applied to $F_\theta$, which arises from the compact support assumption of our probabilistic model. This also means the density function $\nu_\theta$ is quite similar to a mixture of logistics density function $f$. 

Thus we make the following statement:

\begin{lemma}
\label{thm:pnn_mol}
Let $F_\theta$ be a two-layer PNN with hidden dimension $k$ and support bounds $[A, B]$, and $F$ be the cdf of a $k$-mixture of logistics distribution. Then, as $B \rightarrow \infty$, there exists a sequence of parameters $\theta$ for $F_\theta$ so that
\begin{equation}
    |F_\theta(x) - F(x)| < \epsilon
\end{equation}
for all $\epsilon > 0$.
\end{lemma}

Given this connection, we can think of the PNN as parameterizing a "deep" mixture of logistics distribution: the first $n-2$ layers of the neural network transform the input $x$ into a deep hidden representation $a_{n-2} = F_\theta^{n-2}(x)$, and the final two layers of the neural network define a mixture of logistics distribution over this representation $a_{n-2}$. This formulation may explain the representational power of NITS, which we will next explore.

% \subsection{Fast Sampling and Computation of the Jacobian}
% A valid concern may be the runtime of Algorithms \ref{alg:sampling} and \ref{alg:pdf_cdf}. We note that most operations, i.e. the computation of $N_\theta$, $\nu_\theta$, and sampling in the case of coordinate-wise independent random variables, can be parallelized across minibatches and dimensions, with the sole exception being sampling for the autoregressive model.

\subsection{NITS is a Universal Density Estimator}
\label{expressiveness}
In this section, we show that NITS can be used to approximate any continuous, real-valued autoregressive random variable with compact support, given sufficient width and depth of the underlying PNNs.

We start from the one-dimensional case, and show that each conditional distribution $\nu_\theta$ is a universal density estimator.

\begin{theorem}
\label{thm:universal_1d}
(PNN is a universal approximator for positive 1D densities with bounded support)
Let $N_\theta$ be a PNN with weights $\theta$. Suppose $\nu:\mathbb{R} \times \mathbb{R}^{d-1} \rightarrow \mathbb{R}$ is a differentiable one-dimensional conditional density with bounded support, i.e. $\text{supp}(\nu) = [A, B]^d$ for real $A, B$. Then there exists a set of weights $\theta$ such that
\begin{equation}
    |\nu_\theta(x|y) - \nu(x|y)| < \epsilon
\end{equation}
for any $\epsilon > 0$, where $\nu_\theta(x|y) := \nabla_x |_{y=c} N_\theta(x,y)$.
\end{theorem}

Now, noting that any target density $\nu$ of an autoregressive random variable $X = (X_1, \dots X_d)$ can be factored as
\begin{equation}
    \nu(x) = \nu(x_d | x_{d-1},\dots,x_1)\dots\nu(x_1) 
    \label{eq:autoregressive}
\end{equation}
we observe that Theorem \ref{thm:universal_1d} is easily applicable to estimation of $\nu$.

\begin{corollary}
\label{thm:universal}
(PNN is a universal density estimator for general continuous autoregressive random variables) Let $\mu(x)$ be a general joint density for a $d$-dimensional autoregressive random variable, i.e. takes on the form in Eq. \ref{eq:autoregressive}. Then there exists a set of PNNs $\{N_{\theta_i}\}_{i=1}^d$ that induce a $\nu_\theta$ (just as Eq. \ref{eq:autoregressive}) such that for any $\epsilon > 0$,
\begin{equation}
    ||\nu_\theta(x) - \nu(x)||_1 < \epsilon.
\end{equation}
\end{corollary}

Furthermore, the case where $X$ is coordinate-wise independent, i.e.
$
% \begin{equation}
    \mu(x) = \mu(x_d)\dots\mu(x_1),
    \label{eq:independent}
% \end{equation}
$
is also immediately true as a special case of Corollary \ref{thm:universal}, where we assume that $\mu(x_i | x_{i-1}, \dots, x_1) = \mu(x_i)$.

\section{Related Work}
\label{related_work}
While there are many works that take on the task of general density estimation, nearly all approaches avoid direct representation of the pdf itself. The two exceptions are the Real-valued Neural Autoregressive Distribution Estimator (RNADE) \cite{uria2013rnade} and Mixtures of Conditional Gaussian Scale Mixtures (MCGSM) \cite{theis2012mixtures,theis2015generative}, which, like our approach, form an autoregressive assumption on the modeled variable and compute one-dimensional conditional densities over each coordinate of the variable. Notably, like our NITS framework, both RNADE \cite{uria2013rnade} and MCGSM \cite{theis2015generative} employ a network $W_\phi$ to produce parameters for the conditional densities. However, unlike our approach, which approximates the conditional density via a deep neural network, both RNADE and MCGSM use much simpler distributions---mixtures of Gaussians and mixtures of scale Gaussians, respectively.

At large, the general landscape of density estimation models could be understood by how they tackle the partition function $Z_\theta$. In this sense, modern approaches can be categorized into \textbf{two main thrusts}: those that avoid calculating $Z_\theta$ entirely and those that calculate $Z_\theta$ analytically.
%, and those that calculate $Z_\theta$ numerically. 

\textbf{Avoiding $Z_\theta$} \hspace{.05in} Techniques that sidestep calculation of $Z_\theta$ form the brunt of contemporary approaches, and do so via mathematical or architectural constraints. 

Normalizing flows, which were first proposed by \cite{dinh2014nice} for density estimation and extended by many later works, including \cite{rezende2015variational,chen2018neural,kingma2018glow}, define a distribution as the pushforward measure of an invertible transformation $T$, and compute the density via a probabilistic change of variables formula, involving $T^{-1}$ and $dT/dx$, which may be highly expensive, thus requiring the careful design of $T$.

Recently, score-based models \cite{song2019generative,song2020score} have been shown to be highly effective for image generation, and involve estimation of the score function $\nabla_x \log p(x)$ of the model. Crucially, as $Z_\theta$ does not depend on $x$, the gradient with respect to $x$ removes the dependence of the score function on $Z_\theta$. However, this also removes the ability for the model to directly represent the likelihood during training. During inference, \cite{song2020score} show that score-based models can \textit{approximate} $p(x)$ via the pushforward of a normalizing flow. However, the resulting likelihood is not exact, and inherits all the pitfalls of a normalizing flow-based model.

Perhaps the most well-known approach is variational inference. Proposed by \cite{jordan1999introduction}, adapted with stochastic gradient descent by \cite{hoffman2013stochastic}, and applied to deep learning by \cite{kingma2013auto,rezende2014stochastic}, the variational inference framework has achieved remarkable traction in the machine learning community, and involves maximization of a lower bound on the likelihood, called Evidence Lower Bound Optimization (ELBO). As the name suggests, these techniques bound the likelihood rather than directly estimating it.

\textbf{Deriving $Z_\theta$} \hspace{.05in} Much of classical maximum likelihood estimation (MLE) relies on forms of $f_\theta$ for which $Z_\theta$ can be computed analytically. This includes MLE with the exponential family of distributions (e.g. Normal, Gamma, Dirichlet, etc.), the logistic distribution, the von Mises Fisher distribution for hyperspheres, and many more. Of the contemporary approaches, only RNADE, MCGSM, and related works \cite{uria2013rnade,uria2014deep,germain2015made} use analytically tractable partition functions. However, as previously stated, this limits the approaches to simple distributions.

In contrast, our approach can be considered to be in a third category: numerical computation of $Z_\theta$. Due to the computational difficulty of computing high dimensional integrals, we know of no other works in this vein. Note that even our approach has significant drawbacks: our proposed \textit{integration trick} for computing integrals of parametric functions can only compute one-dimensional integrals at a time, which limits our model to an autoregressive architecture.

We also discuss related work involving approximate density models, i.e. probabilistic models that do not support the computation of \textit{exact} model likelihoods, but rather \textit{approximate} model likelihoods. As previously discussed, score-based models can be considered one such approach: while the score function induces a continuous normalizing flow (CNF) whose prior is provably Gaussian when the diffusion time $T$ is infinitely long, the diffusion model is only tractable for finite $T$, for which the prior is intractable. However, \cite{song2020score} still use a Gaussian prior to compute the pushforward measure in the induced CNF. Therefore the pushforward density predicted by the CNF is never exact. Another such approximate density model is the Autoregressive Energy Machine (AEM) proposed in \cite{pmlr-v97-durkan19a}, which (like NITS) also estimate $Z_\theta$ numerically by factorizing the high-dimensional density into autoregressive conditional 1D densities, and estimate the $Z_\theta$ of each 1D density individually. However, while NITS computes $Z_\theta$ exactly, AEMs estimate it via a biased importance sampling scheme. Since both score-based diffusion models and AEMs do not compute exact densities, we do not consider them in the following density estimation benchmarks.

On the broader topic of numerical integration, Neural ODEs \cite{rchen2018neural} and many subsequent works \cite{grathwohl2018ffjord,song2020score} tackle the related task of integrating an arbitrary 1D function over some interval $[t_0, t_1]$. However, rather than estimating the desired function as a derivative of another differentiable function, these works model the desired function directly, and use a differentiable ODE solver to perform integrations.

Additionally, though the idea of representing monotone functions using neural networks has been thoroughly explored \cite{archer1993application,sill1998monotonic,daniels2010monotone,gupta2016monotonic}, our application to \textit{direct} density estimation is novel. \cite{chen2018neural} explore the use of a monotone neural network function to augment the autoregressive transformation in Inverse Autoregressive Flows (IAF) \cite{kingma2016improved}, which in the original formulation is a simple linear transform. This was further extended by \cite{wehenkel2019unconstrained}, who compute monotone functions by integrating positive neural network functions with an ODE solver.

\begin{figure}
\vskip 0.2in
\begin{center}
% \centerline{\includegraphics[width=\columnwidth]{pixelcnn_ours}}
\centering
\subfigure{\includegraphics[width=75mm]{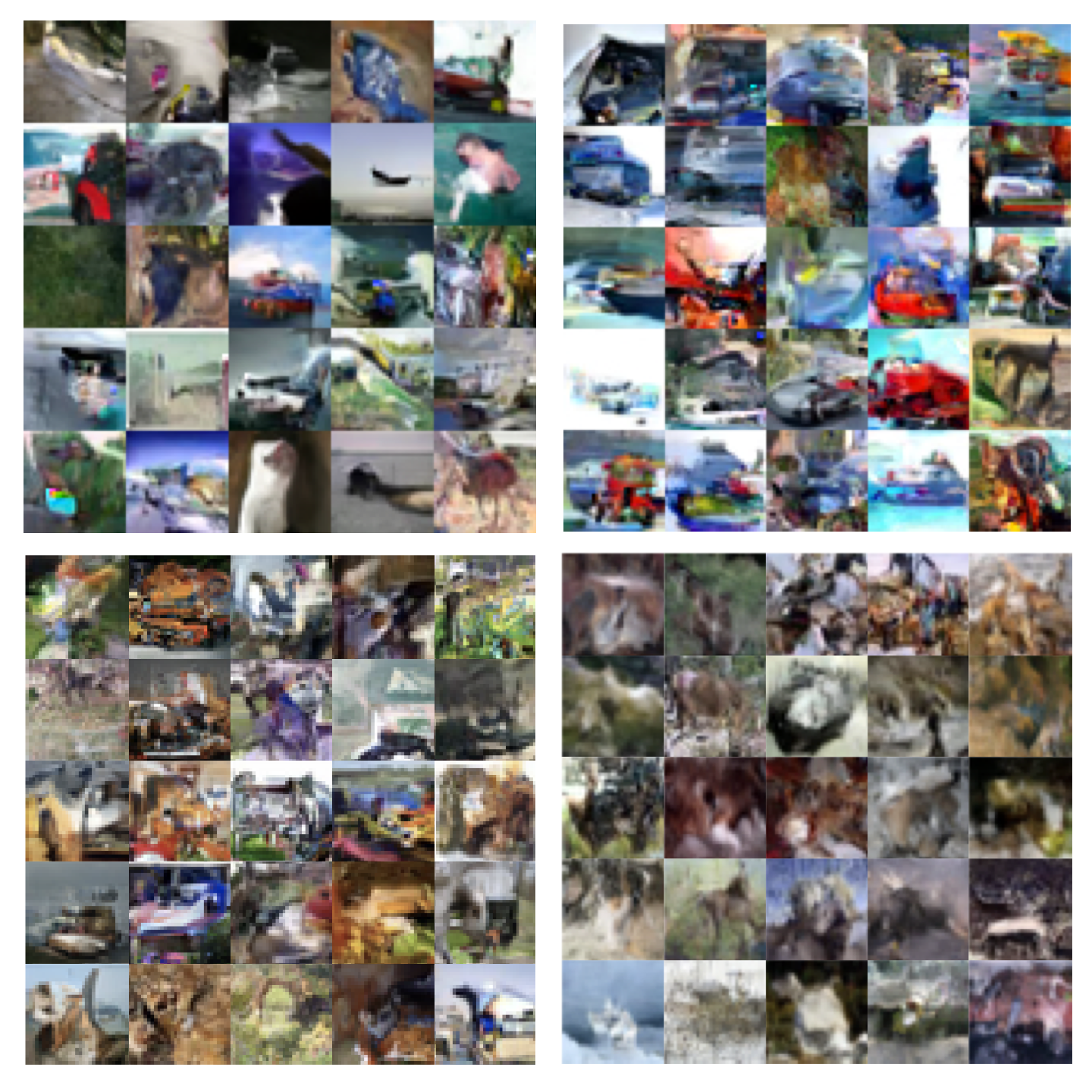}}

\caption{Randomly generated images from DISCRETE NITS-CONV (top left) and NITS-CONV (top right). Compare with competing discretized and continuous density models, Pixel CNN (bottom left) and Flow++ (bottom right), respectively. }
\label{fig:cifar10}
\end{center}
\vskip -0.2in
\end{figure}

\begin{table}
\caption{Negative log likelihood (in bits/dim) for CIFAR-10. The table is split into halves, with discretized density models above and continuous density models below. We obtain competitive results among both types of models.}
\label{tab:cifar10}
\vskip 0.15in
\begin{center}
\begin{small}
\begin{sc}
\begin{tabular}{lcr}
\toprule
Model & CIFAR-10 \\
\midrule
Pixel CNN & 3.14 \\
Gated Pixel CNN & 3.03 \\
Row Pixel RNN & 3.00 \\
Pixel CNN++ & 2.92 \\
Image Transformer & 2.90 \\
PixelSNAIL & 2.85 \\
DISCRETE NITS-CONV (ours) & 2.94 \\
\midrule
RealNVP & 3.49 \\
Glow & 3.35 \\
% NSF &
Flow++ & 3.08 \\
NITS-CONV (ours) & 2.97 \\
\bottomrule
\end{tabular}
\end{sc}
\end{small}
\end{center}
\vskip -0.1in
\end{table}

\begin{table*}[t]
\caption{Test log likelihood for UCI datasets and BSDS300, with error bars corresponding to two standard deviations. The table is split into two halves: the upper half denotes flow-based models, and the lower half denotes autoregressive continuous density models. NITS-CONV is only applied to BSDS300, as the convolutional architecture is only readily applicable to images.}
\label{tab:uci}
\vskip 0.15in
\begin{center}
\begin{small}
\begin{sc}
\begin{tabular}{lccccr}
\toprule
Model & POWER & GAS & HEPMASS & MINIBOONE & BSDS300 \\
\midrule
MAF                 & 0.30 $\pm$ 0.01 & 9.59 $\pm$ 0.02 & -17.39 $\pm$ 0.02 & -11.68 $\pm$ 0.44 & 156.36 $\pm$ 0.28\\
TAN                 & 0.48 $\pm$ 0.01 & 11.19 $\pm$ 0.02 & -15.12 $\pm$ 0.02 & -11.01 $\pm$ 0.48 & 157.03 $\pm$ 0.07\\
NAF                 & 0.62 $\pm$ 0.02 & 11.91 $\pm$ 0.13 & -15.09 $\pm$ 0.40 & \textbf{-8.86 $\pm$ 0.15} & 157.73 $\pm$ 0.04\\
B-NAF               & 0.61 $\pm$ 0.01 & 12.06 $\pm$ 0.02 & -14.71 $\pm$ 0.02 & -8.95 $\pm$ 0.07 & 157.36 $\pm$ 0.03\\
FFJORD              & 0.46 $\pm$ 0.01 & 8.59 $\pm$ 0.12 & -14.92 $\pm$ 0.08 & -10.43 $\pm$ 0.04 & 157.40 $\pm$ 0.19\\
SOS                 & 0.60 $\pm$ 0.01 & 11.99 $\pm$ 0.41 & -15.15 $\pm$ 0.10 & -8.90 $\pm$ 0.11  & 157.48 $\pm$ 0.41\\
NSF                 & \textbf{0.66 $\pm$ 0.01} & 13.09 $\pm$ 0.02 & -14.01 $\pm$ 0.03 & -9.22 $\pm$ 0.48 & 157.31 $\pm$ 0.28\\
RealNVP             & 0.17 $\pm$ 0.01 & 8.33 $\pm$ 0.14 & -18.71 $\pm$ 0.02 & -13.84 $\pm$ 0.52 & 153.28 $\pm$ 1.78\\
\midrule
MADE MoG            & 0.40 $\pm$ 0.01 & 8.47 $\pm$ 0.02 & -15.15 $\pm$ 0.02 & -12.27 $\pm$ 0.47 & 153.71 $\pm$ 0.28 \\
NITS-MLP (ours)         & \textbf{0.66 $\pm$ 0.01}   & \textbf{13.20 $\pm$ 0.01}  & \textbf{-12.93 $\pm$ 0.02} & -10.85 $\pm$ 0.02  & 155.91 $\pm$ 0.21\\
NITS-CONV (ours)    & -                 & -                 & -                 & -                  & \textbf{163.35 $\pm$ 0.22}\\
\bottomrule
\end{tabular}
\end{sc}
\end{small}
\end{center}
\vskip -0.1in
\end{table*}

\section{Experiments}
In this section, we extensively evaluate the capabilities of our proposed approach via density estimation on two vastly different modalities. In our first experiment, we propose two NITS-based image density models, based on the causal convolution architecture proposed in \cite{van2016pixel} and refined in \cite{salimans2017pixelcnn++,chen2018pixelsnail}. We evaluate our two models on the CIFAR-10 dataset, where we demonstrate the state-of-the-art performance among continuous density estimators (Section \ref{exp:pixelcnn}). In the second experiment, we benchmark NITS against competing density estimation models with a suite of UCI datasets (Section \ref{exp:uci}).

\subsection{Pixel-wise Density Estimation with PixelCNN++}
\label{exp:pixelcnn}
First, we evaluate our approach against other generative image density estimation models on the CIFAR-10 dataset. To improve the capacity of our model, we use a convolutional neural architecture in our weight model $W_\phi$, employing causally masked \cite{van2016pixel} atrous \cite{sermanet2013overfeat} convolutions with residual \cite{he2016deep} and short-cut \cite{salimans2017pixelcnn++} connections. As the usage of a convolutional weight model still follows the original formulation of NITS, we refer to it simply by the weight model architecture, i.e. NITS-CONV.

Additionally, we explore the performance of NITS-CONV when exploiting the 8-bit quantization of the CIFAR-10 dataset. For this approach, we apply a similar discretization procedure as in \cite{salimans2017pixelcnn++}, where the likelihood of a given intensity value of a given pixel $x_i$ is given by
\begin{equation*}
    pmf(x_i = a) = \int_A^{\lceil a \rceil} pdf(t) dt - \int_A^{\lfloor a \rfloor} pdf(t) dt,
\end{equation*}
for $a \in \{0, 1, 2, 3, \dots, 255\}$. The integral computation on the right hand side is quite simple for our NITS density model, due to fast integration via the \textit{integration trick}. We differentiate this variant of our model from the original with the \textit{discretized} modifier, i.e. DISCRETE NITS-CONV.

One notable novelty of NITS-CONV is that it is the first modern autoregressive image density model with a continuous density model. Thus the parameterization of our model does not rely on the quantization factor
of the modeled images. This bridges the gap between autoregressive density models and flow-based models, which are all continuous density models.

In Table \ref{tab:cifar10}, we compare our model to both discretized density models (i.e., PixelCNN \cite{van2016pixel} and its variants \cite{salimans2017pixelcnn++,chen2018pixelsnail,oord2016conditional}, and Image Transformer \cite{parmar2018image}) and continuous density models (i.e. RealNVP \cite{dinh2016density}, Glow\cite{kingma2018glow}, and Flow++ \cite{ho2019flow++}). 
% Among continuous density models, we report the best negative log-likelihood scores in bits/dim. 
See Figure \ref{fig:cifar10} for examples of generated images.

\subsection{Density Estimation on UCI Datasets}
\label{exp:uci}
We additionally evaluate our method against the UCI suite of density estimation benchmarks. We compare our approach against Masked Autoregressive Flows (MAF) \cite{papamakarios2017masked}, Transformation Autoregressive Networks (TAN) \cite{oliva2018transformation}, Neural and Block-Neural Autoregressive Flows (NAF and B-NAF) \cite{chen2018neural,de2020block}, Free-form Jacobian of Reversible Dynamics (FFJORD) \cite{grathwohl2018ffjord}, Sum of Squares Polynomial Flow (SOS) \cite{jaini2019sum}, RealNVP \cite{dinh2016density}, and Masked Autoencoder Density Estimation \cite{germain2015made}. The results are shown in Table \ref{tab:uci}. 

We report state of the art density estimation on four out of five benchmarks in the UCI and BSDS300 datasets. Furthermore, our continuous density autoregressive model outperforms the only other competing continuous density autoregressive model that has been historically applied to the UCI datasets, MADE \cite{germain2015made}, on all datasets.

\section{Conclusion}
In this work we introduced two novel computational ideas: fast integration over arbitrary densities via the \textit{integration trick}, and fast sampling from arbitrary densities via the inverse transform method. We combine these ideas in the design of the Neural Inverse Transform Sampler (NITS), which enjoys high expressiveness, fast density estimation, fast sampling, and end-to-end differentiability. Notably, NITS is (to our knowledge) the first model to provide explicit density approximation for densities with non-analytical partition functions. We benchmark NITS against existing density estimators, and show that NITS exhibits state of the art performance on several datasets.

% Note use of \abovespace and \belowspace to get reasonable spacing
% above and below tabular lines.

% % Acknowledgements should only appear in the accepted version.
% \section*{Acknowledgements}

% \textbf{Do not} include acknowledgements in the initial version of
% the paper submitted for blind review.

% If a paper is accepted, the final camera-ready version can (and
% probably should) include acknowledgements. In this case, please
% place such acknowledgements in an unnumbered section at the
% end of the paper. Typically, this will include thanks to reviewers
% who gave useful comments, to colleagues who contributed to the ideas,
% and to funding agencies and corporate sponsors that provided financial


\bibliography{main}
\bibliographystyle{icml2020}

\newpage
\clearpage

\appendix
% \section{Do \emph{not} have an appendix here}

% \textbf{\emph{Do not put content after the references.}}
% %
% Put anything that you might normally include after the references in a separate
% supplementary file.

% We recommend that you build supplementary material in a separate document.
% If you must create one PDF and cut it up, please be careful to use a tool that
% doesn't alter the margins, and that doesn't aggressively rewrite the PDF file.
% pdftk usually works fine. 

% \textbf{Please do not use Apple's preview to cut off supplementary material.} In
% previous years it has altered margins, and created headaches at the camera-ready
% stage. 
% %%%%%%%%%%%%%%%%%%%%%%%%%%%%%%%%%%%%%%%%%%%%%%%%%%%%%%%%%%%%%%%%%%%%%%%%%%%%%%%
% %%%%%%%%%%%%%%%%%%%%%%%%%%%%%%%%%%%%%%%%%%%%%%%%%%%%%%%%%%%%%%%%%%%%%%%%%%%%%%%

\section{Proofs}

\subsection{PNNs Compute Valid Density Functions}
\begin{proof}
(Lemma \ref{thm:monotonic}) Let $F_\theta$ be an $\ell$-layer neural network with positive weights $W$ and positive monotonic activations $\sigma$ of the general form defined by the recursive equation
\begin{equation*}
    a_i = \sigma(W_i^T a_{i-1} + b_i) \hspace{.05in} \text{for $i=1,\dots,n$},
\end{equation*}
where $a_i$ is the activation of the $i$-th layer of $F_\theta$, $a_\ell = F_\theta(x)$, and $a_0(x) = x$. We note that
\begin{align*}
    \frac{d a_i}{d a_{i-1}} &= (W_i \sigma'(W_i^T a_{i-1} + b_i))^T,
\end{align*}
where, since $(W_i)_{jk} > 0$ for all $j,k$ and $\sigma'(W_i^T a_{i-1} + b_i)_j > 0$ for all $j$ by assumption, $\frac{d a_i}{d a_{i-1}} > 0$.

Thus, by the chain rule of differentiation,
\begin{align*}
    \nabla_x F_\theta(x) 
    &= \frac{d a_\ell}{d a_{\ell-1}} \frac{d a_{\ell-1}}{d a_{\ell-2}}  \dots \frac{d a_2}{d a_1} \\
    &> 0.
\end{align*}
Since this argument is independent of $x$, we have shown the statement for all $x$.
\end{proof}

\subsection{NITS is a Mixture of Logistics Distribution}
\begin{proof}
(Lemma \ref{thm:pnn_mol}) According to Eq. \ref{eq:mol_like_cdf}, the final two layers of the PNN can be written as
\begin{equation*}
    N_\theta(x) = Z_\theta \sum_{i=1}^k \beta_i \sigma((x - b_i) / s_i)
\end{equation*}
where $Z_\theta = \int \nabla_x F_\theta(x) dx$ is the partition function. Further note that since $N_\theta$ is the scaled convex combination of sigmoid functions, $Z_\theta \rightarrow 1$ as $A$ and $B$ tend toward positive and negative infinity, respectively. Therefore this function is equivalent to the cdf of a mixture of logistics function (i.e. Eq. \ref{eq:mol_cdf}) in the limit.
\end{proof}

\subsection{NITS is a Universal Density Estimator}
We first show the following lemma:
\begin{lemma}
\label{thm:universal_cdf}
Let $N_\theta$ be a PNN with compact boundary $[A^*, B^*]$ and $N$ be an arbitrary positive monotonic cdf with bounded density support $[A, B]$. Then for any $\epsilon > 0$ there exists an $A^*, B^*$ and set of parameters $\theta$ such that
\begin{equation}
    || N_\theta(x) - N(x) ||_{L_\infty([A, B])} < \epsilon.
\end{equation}
\end{lemma}

\begin{proof}
(Lemma \ref{thm:universal_cdf})
Choose $\epsilon_1, \epsilon_2$ so that $\epsilon = \epsilon_1 + \epsilon_2$. Define the cumulative distribution function (cdf) corresponding to the density $\nu$ as
\begin{equation}
    N(x) := \int_\mathbb{R} \nu(x) dx = \int_A^B \nu(x) dx.
\end{equation}

We shall use Lemma 2 in \cite{chen2018neural}, where it is shown that, given $\epsilon_1 > 0$ and an arbitrary positive monotonic function $N: \mathbb{R} \rightarrow [0, 1]$, there exists a $\theta = (\{\mu_i\}_{i=1}^n, \{s_i\}_{i=1}^n)$ such that a function of the form
\begin{equation}
    F(x) = \sum_{i=1}^n \alpha_i \sigma\left(\frac{x - \mu_i}{s_i}\right)
\end{equation}
will be $\epsilon_1$-close to $N$ in $L_\infty([A,B])$, i.e. 
\begin{equation}
    |F(x) - N(x)| < \epsilon_1 \hspace{.1in} \text{for all $x \in [A, B]$}
\end{equation}

A one layer PNN with bounds $[A', B']$ can be written in the form
\begin{equation}
    N_\theta(x) = \frac{F(x)}{F(B') - F(A')}.
\end{equation}

Letting $C_{A', B'} = \frac{1}{F(B') - F(A')}$, and noting that $F(x) < 1$ for all $x \in \mathbb{R}$, we have
\begin{align*}
    |F(x)(1 - C_{A', B'})| 
    &\leq |1 - C_{A', B'}| \\
    &= \frac{1}{F(B') - F(A')} - 1.
\end{align*}
Now we pick $A^*, B^*$ so that
\begin{equation}
    F(B^*) - F(A^*) > \frac{1}{\epsilon_2 + 1}.
\end{equation}
The existence of such a $A^*, B^* \in \mathbb{R}$ is guaranteed by the fact that $\frac{1}{\epsilon_2 + 1} < 1$ and $\lim_{x \rightarrow \infty} F(x) = 1$ and $\lim_{x \rightarrow -\infty} F(x) = 0$.
Therefore,
\begin{equation}
    |F(x)(1 - C_{A^*, B^*})| < \epsilon_2.
\end{equation}

And so by choosing the bounds of the PNN $N_\theta$ as $(A^*, B^*)$,
\begin{align*}
    |N_\theta(x) - N(x)| 
    &= |CF(x) - N(x) + F(x)(1 -  \\
    & \hspace{.25in} C_{A^*, B^*}) - F(x)(1 - C_{A^*, B^*})| \\
    &\leq |F(x) - N(X)| + |F(x)(1 - \\
    & \hspace{.25in} C_{A^*, B^*})| \\
    &\leq \epsilon_1 + \epsilon_2 = \epsilon.
\end{align*}

\end{proof}

Now we show the main theorem.

\begin{proof}
(Theorem \ref{thm:universal_1d})
Choose $\epsilon_3, \epsilon_4$ so that $\epsilon = \epsilon_3 + \epsilon_4$. As $N_\theta$ and $\nu$ are both differentiable, we define
\begin{align}
    A_h &= \frac{N_\theta(x + h) - N_\theta(x)}{h} \\
    B_h &= \frac{N(x + h) - N(x)}{h},
\end{align}
and observe that for any $\epsilon_3$, there exists $h$ such that
\begin{equation*}
    |A_h - \nabla_x N_\theta(x)| < \epsilon_3/2
    |B_h - \nabla_x N(x)| < \epsilon_3/2.
\end{equation*}
Moreover, from Lemma $\ref{thm:universal_cdf}$, we observe that there exists some $\theta$ and bounds $[A^*, B^*]$ such that 
\begin{align*}
    |A_h - B_h| 
    &= \left|\frac{N_\theta(x + h) - N_\theta(x)}{h} - \frac{N(x + h) - N(x)}{h} \right|\\
    &= \frac{\left|N_\theta(x + h) - N(x + h) - N_\theta(x) - N(x) \right|}{h}\\
    &\leq \frac{|N_\theta(x + h) - N(x + h)| + |N_\theta(x) - N(x)|}{h}\\
    &\leq \epsilon_4.
\end{align*}

Thus,
\begin{align*}
    &|\nabla_x N_\theta(x) - \nabla_x N(x)| \\
    &= \left|\nabla_x N_\theta(x) - \nabla_x N(x) + (A_h - B_h) - (A_h - B_h) \right| \\
    &\leq |\nabla_x N_\theta(x) - A_h| + |\nabla_x N(x) - B_h| + |A_h - B_h| \\
    &\leq \epsilon_3 + \epsilon_4 = \epsilon
\end{align*}

\end{proof}

Finally, we show the autoregressive corollary.

\begin{proof}
(Corollary \ref{thm:universal})
Briefly, the sketch of the proof involves bounding each joint density
\begin{equation*}
    ||\nu_\theta(x_i, \dots, x_1) - \nu_\theta(x_i, \dots, x_1)|| < \epsilon_i,
\end{equation*}
by some $\epsilon_i$ for all $i$, such that some recursive inequality of $||\nu_\theta(x) - \nu(x)||_1$, subsequently defined, is ultimately bounded by $\epsilon$.

We first observe the following inequality:
\begin{align*}
    ||ab - cd|| 
    &= \frac{1}{2} ||(a + c)(b - d) + (a - c)(b + d)|| \\
    &\leq \frac{1}{2} \left( ||a + c|| \, ||b - d|| + ||a - c|| \, ||b + d|| \right).
\end{align*}

Now, we note that for each $i = 1, \dots, n$, we may choose
\begin{align*}
    a &= \nu_\theta(x_i|x_{<i}), \hspace{.2in} & b = \nu_\theta(x_{<i}), \\
    c &= \nu(x_i | x_{<i}), \hspace{.2in} & d = \nu(x_{<i}),
\end{align*}
where we define $\nu_\theta(x_0) = \nu(x_0) = 1$.

Applying the above inequality, we can see that
\begin{align*}
    &||\nu_\theta(x_i|x_{<i}) \nu_\theta(x_{<i}) - \nu(x_2 | x_{<i}) \nu(x_{<i})||_1 \\
    &\leq \frac{1}{2} \left(C_{i1} ||a - c||_1 + C_{i2} ||b - d||_1 \right), \numberthis \label{eq:recursive_def}
\end{align*}
where $C_1 = \nu_\theta(x_i|x_{<i}) + \nu(x_i | x_{<i})$ and $C_2 = \nu_\theta(x_{<i}) + \nu(x_{<i})$ are bounded quantities due to the boundedness of $\nu_\theta$ and $\nu$, and $||a - c||_1 = ||\nu_\theta(x_i|x_{<i}) - \nu(x_i | x_{<i})||_1$ and $||b - d||_1 = ||\nu_\theta(x_{<i}) - \nu(x_{<i})||$. Notice that this is a recursive definition, as $||b - d||_1$ is simply Eq. \ref{eq:recursive_def} at $i - 1$.

Therefore, if we choose
\begin{equation*}
    \epsilon_i = \frac{\epsilon}{\prod_{j=i+1}^n C_{j2} C_{i1} 2^{i-2}},
\end{equation*}
we can see that
\begin{align*}
     ||\nu_\theta(x) - \nu(x)||_1
     &\leq \sum_{i=1}^n \frac{\epsilon}{2^i} \\
     &< \epsilon
\end{align*}

% For $i=1$, we know by Theorem \ref{thm:universal_1d} that there exists parameters $\{\theta, A, B\}$ such that
% \begin{equation}
%     ||\nu_\theta(x_1) - \nu(x_1)||_1 \leq \epsilon_1.
% \end{equation}
% Our choice of $\epsilon_1 < \epsilon$ will be discussed later.

% Now, before we show the inductive step, we observe the following inequality:
% \begin{align*}
%     ||ab - cd|| 
%     &= \frac{1}{2} ||(a + c)(b - d) + (a - c)(b + d)|| \\
%     &\leq \frac{1}{2} \left( ||a + c|| \, ||b - d|| + ||a - c|| \, ||b + d|| \right).
% \end{align*}
% For the inductive step: at time $i > 1$, we let
% \begin{align*}
%     a &= \nu_\theta(x_i|x_{<i}), \hspace{.2in} & b = \nu_\theta(x_{<i}), \\
%     c &= \nu(x_i | x_{<i}), \hspace{.2in} & d = \nu(x_{<i}).
% \end{align*}
% Applying the above inequality, we can see that
% \begin{align*}
%     &||\nu_\theta(x_i|x_{<i}) \nu_\theta(x_{<i}) - \nu(x_2 | x_{<i}) \nu(x_{<i})||_1 \\
%     &\leq \frac{1}{2} \left(C_1 ||a - c||_1 + C_2 ||b - d|| \right),
% \end{align*}
% where $C_1 = \nu_\theta(x_i|x_{<i}) + \nu(x_i | x_{<i})$ and $C_2 = \nu_\theta(x_{<i}) + \nu(x_{<i})$ are bounded quantities due to the boundedness of $\nu_\theta$ and $\nu$. Thus we can pick 
\end{proof}

\section{Experimental Hyperparameters}
The architecture used by NITS-MLP, which has a fully connected weight model with residual connections as used in \cite{pmlr-v97-durkan19a}. More details are included in Table \ref{tab:nits_fc}. Models were trained with early stopping with a patience of 5 epochs, and a learning rate of $2e-4$.

For NITS-CONV, which has a convolutional weight model, the architecture involves a block ResNet architecture involving causally masked atrous convolutional layers, as described in \ref{exp:pixelcnn}. The model applied to the CIFAR-10 dataset is comprised of 5 ResNet blocks, and contains a total of 53M parameters. The model applied to the BSDS300 dataset is comprised of a single ResNet block, and contains a total of 3.8M parameters. Models were trained for 150 epochs with a learning rate of $2e-4$

All PNN architectures involved two hidden layers with 16 hidden units. All models were trained on a Nvidia RTX 2080 Ti GPU.

\begin{table*}[t]
\caption{Information and hyperparameters for NITS-MLP on UCI Datasets.}
\label{tab:nits_fc}
\vskip 0.15in
\begin{center}
\begin{small}
\begin{sc}
\begin{tabular}{lcccccr}
\toprule
& POWER & GAS & HEPMASS & MINIBOONE & BSDS300 \\
\midrule
Dimension & 6 & 8 & 21 & 43 & 63 \\
Train Set Size & 1,615,917 & 852,174 & 315,123 & 29,556 & 1,000,000 \\
\midrule
Batch Size & 1024 & 1024 & 1024 & 128 & 1024 \\
Hidden Dim & 1024 & 1024 & 512 & 128 & 1024 \\
Dropout & 0.2 & 0.2 & 0.3 & 0.1 & 0.2 \\
Residual Blocks & 4 & 4 & 4 & 8 & 4 \\
% $2 \times 256$ & $2 \times 256$ & $2 \times 256$ & $2 \times 128$ & $2 \times 1024$ \\
\bottomrule
\end{tabular}
\end{sc}
\end{small}
\end{center}
\vskip -0.1in
\end{table*}

\begin{figure*}[ht]
\vskip 0.2in
\begin{center}
\subfigure{%
\label{fig:first}%
\includegraphics[height=3in]{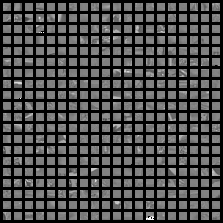}}%
\hspace{.15in}
\subfigure{%
\label{fig:second}%
\includegraphics[height=3in]{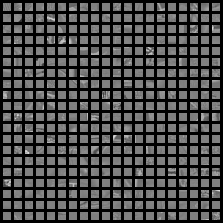}}%

\caption{Additional generated images from NITS-CONV trained on the BSDS300 dataset (left) and true images from the dataset (right).}
\label{fig:additional_bsds}
\end{center}
\vskip -0.2in
\end{figure*}

\end{document}